\documentclass[10pt,twocolumn,letterpaper]{article}

\usepackage{iccv}
\usepackage{times}
\usepackage{epsfig}
\usepackage{graphicx}
\usepackage{amsmath}
\usepackage{amssymb}
\usepackage{color}

\usepackage[pagebackref=true,breaklinks=true,letterpaper=true,colorlinks,bookmarks=false]{hyperref}

 \iccvfinalcopy % *** Uncomment this line for the final submission

\def\iccvPaperID{771} % *** Enter the ICCV Paper ID here

\ificcvfinal\fi

\begin{document}

%%%%%%%%% TITLE
\title{DeepBox:  Learning Objectness with Convolutional Networks}

\author{Weicheng Kuo \qquad Bharath Hariharan \qquad Jitendra Malik\\
University of California, Berkeley\\
{\tt\small \{wckuo, bharath2, malik\}@eecs.berkeley.edu}
}

\maketitle
%\thispagestyle{empty}

%%%%%%%%% ABSTRACT
\begin{abstract}
Existing object proposal approaches use primarily bottom-up cues to
rank proposals, while we believe that ``objectness\char`\"{} is in
fact a high level construct. We argue for a data-driven, semantic approach
for ranking object proposals. Our framework, which we call DeepBox, uses
convolutional neural networks (CNNs) to rerank proposals from a bottom-up method. 
We use a novel four-layer CNN architecture that is as good as much larger networks on the task of evaluating objectness while being much faster.
We show that DeepBox significantly improves over the bottom-up ranking, achieving the same recall with 500 proposals as achieved by bottom-up methods with 2000. This improvement generalizes to categories the CNN has never seen before and leads to a 4.5-point gain in detection mAP. Our implementation achieves this performance while running at 260 ms per image.

% ~\textcolor{red}{Finally, we implement multiscale fast DeepBox and achieve $44\%$ improvement over Edge boxes on COCO running at $0.26$s per image on top of Edge boxes using single thread.} 

\end{abstract}

%%%%%%%%% BODY TEXT
\section{Introduction}

Object detection methods have moved from scanning window approaches~\cite{FelzenszwalbPAMI2010}
to ones based on bottom-up object proposals~\cite{GirshickCVPR2014}.
Bottom-up proposals~\cite{AlexeTPAMI2012} have two major advantages: 1) by reducing the search space, they allow the usage of more sophisticated recognition machinery, and 2) by pruning away false positives, they make detection easier. ~\cite{HosangArxiv2015,FastRCNN}

Most object proposal methods rely on simple
bottom-up grouping and saliency cues. The rationale for this is that this step should be
reasonably fast and generally applicable to all object categories.
However, we believe that there is more to objectness than bottom-up grouping or saliency. For instance, many disparate object categories might share high-level structures (such as the limbs of animals and robots) and detecting such structures might hint towards the presence of objects. A proposal method that incorporates these and other such cues is likely to perform much better. 

\begin{figure}
\begin{center}
\includegraphics[width=1\columnwidth]{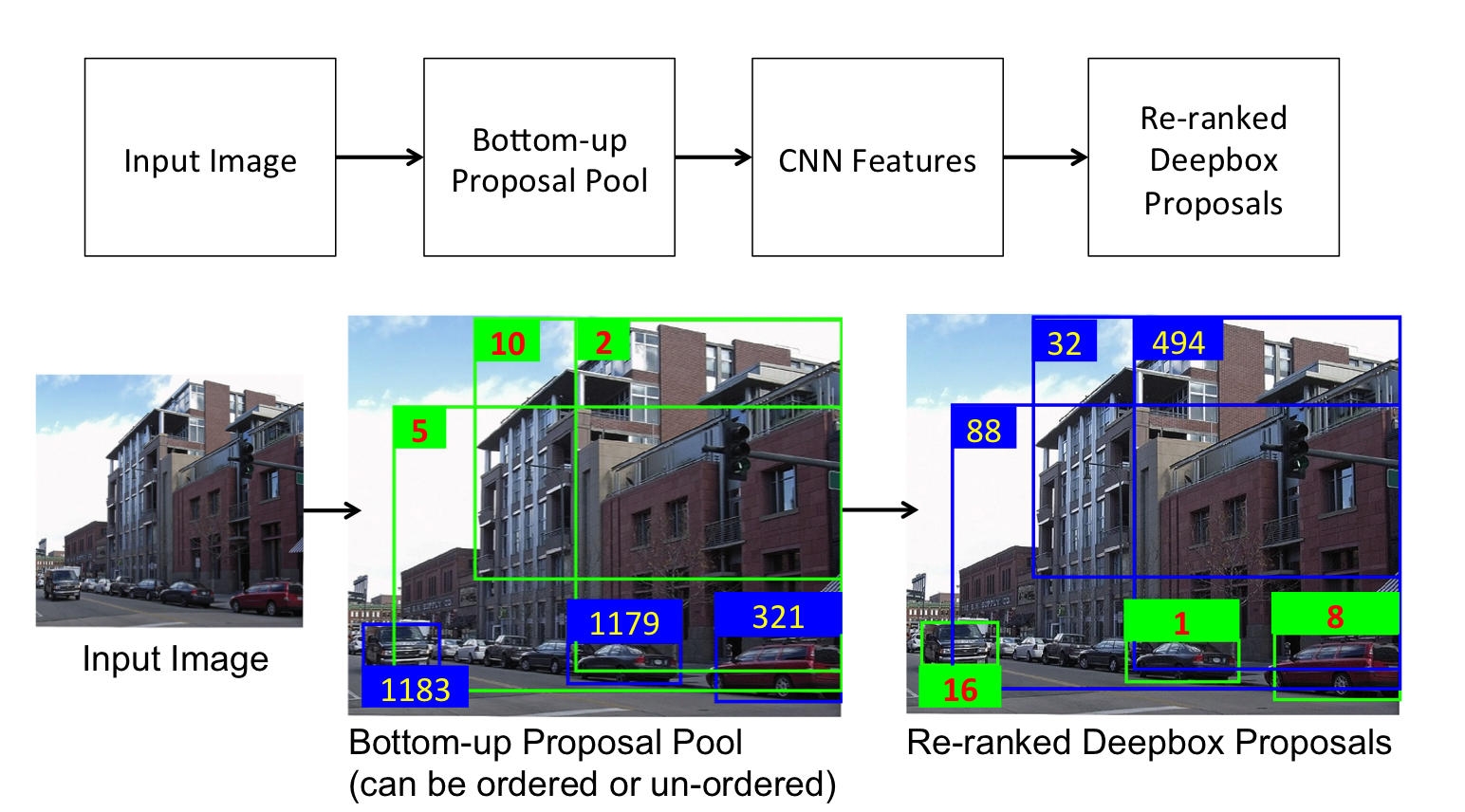}
\end{center}
\caption{The DeepBox framework. Given any RGB image, we first generate bottom-up proposals and then rerank them using  a CNN. High ranked boxes are shown in green and low ranked ones are blue. The number in each box is its ranking in the proposal pool. DeepBox corrects the ranking of Edge box, ranking objects higher than background.}
\label{fig:intro}
\end{figure}

In this paper, we argue for a \emph{semantic}, \emph{data-driven} notion of objectness. 
Our approach is to present a large database of images with annotated objects to a learning algorithm, and let the algorithm figure out what low-, mid- and high-level cues are most discriminative of objects. Following recent work on a range of recognition
tasks~\cite{GirshickCVPR2014,BharathECCV2014,KrizhevskyNIPS2012},
we use convolutional networks (CNNs)~\cite{lecun1989backpropagation} for this task. Concretely, we train a CNN to rerank a large pool of object proposals produced by a bottom-up proposal method (we use Edge boxes~\cite{ZitnickECCV2014} for most experiments in this paper). For ease of reference, we call our approach DeepBox. Figure~\ref{fig:intro} shows our framework.

We propose a lightweight four layer network architecture that significantly improves over bottom-up proposal methods in terms of ranking (26\% relative improvement on AUC over Edge boxes on VOC 2007~\cite{pascal-voc-2007}). Our network architecture is as effective as state-of-the-art classification networks on this task while being much smaller and thus much faster. In addition, using ideas from SPP~\cite{HeECCV2014} and Fast R-CNN~\cite{FastRCNN}, our implementation runs in 260 ms per image, comparable to some of the fastest bottom-up proposal approaches like Edge Boxes (250 ms). We also provide evidence that what our network learns is category-agnostic:  our improvements in performance generalize to categories that the CNN has not seen before (16\% improvement over Edge boxes on COCO~\cite{mscoco}). Our results suggest that a) there is indeed a generic, semantic notion of objectness beyond bottom-up saliency, and that b) this semantic notion of objectness can be learnt effectively by a lightweight CNN.

%Finally, we implemented a fast version of our system using ideas from~\cite{SPP} and~\cite{FastRCNN}. This fast version of Deep Box runs in about 260 milliseconds per image while retaining the same high performance.
%~\textcolor{red}{Finally, to speed up the system further, we implement multiscale Fast DeepBox and achieve $44\%$ AUC improvement over Edge boxes at IoU=0.7 on COCO running at $0.26$s per image on top of Edge boxes. The combined runtime of DeepBox is $0.5$s, which is faster than any segment-based methods due to our use of a light-weight network.}

Object proposals are just the first step in an object detection system, and the final evaluation of a proposal system is the impact it has on detection performance. We show that the Fast R-CNN detection system~\cite{FastRCNN}, using 500 DeepBox proposals per image, is 4.5 points better than the same object detector using 500 Edge box proposals. Thus our high quality proposals directly lead to better object detection.% compared results onused our (fast) Deep Box proposals to train and test Fast-RCNN ~\cite{FastRCNN} object detection experiments on the proposals produced by Fast DeepBox and Edge boxes~\cite{ZitnickECCV2014}. The results show that TBD.} 

The rest of the paper is laid out as follows. In Section~\ref{sec:relwork} we discuss related work. We describe our network architecture and training and testing procedures in Section~\ref{sec:method}. Section~\ref{sec:experiments} describes experiments and we end with a discussion.

%-------------------------------------------------------------------------

\section{Related work}
\label{sec:relwork}
Russell et al.~\cite{RussellCVPR2006} were one of the first to suggest a category-independent method to propose putative
objects. Their method involved sampling regions from multiple segmentations
of the image. More recently, Alexe et al.~\cite{AlexeTPAMI2012} and Endres et al.~\cite{endres2010category} propose using bottom-up object proposals as a first step in recognition. Expanding on the multiple segmentations idea, Selective search~\cite{UijlingsIJCV2013} uses regions from hierarchical segmentations in multiple color
spaces as object proposals.
CPMC~\cite{CarreiraTPAMI2012}
uses multiple graph-cut based segmentations with multiple foreground
seeds and multiple foreground biases to propose objects. GOP~\cite{KrahenbuhlECCV2014} replaces graph cuts with a much
faster geodesic based segmentation. MCG~\cite{ArbelaezCVPR2014}
also uses multiple hierarchical segmentations from different scales
of the image, but produces proposals by combinatorially grouping regions.
Edge boxes~\cite{ZitnickECCV2014} uses contour information instead
of segments: bounding boxes which have fewer contours straggling the
boundary of the box are considered more likely to be objects. 

Many object proposal methods also include a ranking of the regions.
This ranking is typically based on low level region features such
as saliency~\cite{AlexeTPAMI2012}, and is sometimes learnt~\cite{ArbelaezCVPR2014,CarreiraTPAMI2012}.
Relatively simple ranking suffices when the goal is a few thousand proposals as in 
MCG~\cite{ArbelaezCVPR2014}, but to narrow the list down to a few hundred
as in CPMC~\cite{CarreiraTPAMI2012} requires more involved reasoning. 
DeepBox aims at such a ranking.

%The sophistication of the ranking used depends on the regime one wants
%to operate in: MCG~\cite{ArbelaezCVPR2014} for instance optimizes
%for the case when one wants a couple of thousand proposals, while
%CPMC~\cite{CarreiraTPAMI2012} optimizes for a few hundred proposals.
%The latter case requires more complex ranking functions. Ranking is
%an essential part of our DeepBox framework. 

Multibox~\cite{ErhanCVPR2014,SzegedyArxiv2014} directly produces object proposals from images using a sophisticated neural network. In contemporary work, Faster R-CNN~\cite{FasterRCNN} uses the same large network to propose objects and classify them. DeepMask~\cite{Deepmask15} also uses a very deep network to directly produce segment proposals. In comparison, our architecture is quite lightweight and can be used out of the box to rerank any bottom-up proposals.

Finally, we direct the reader to~\cite{HosangBMVC2014,HosangArxiv2015} for a more thorough evaluation of bottom-up proposal methods.
\section{Method}
\label{sec:method}
The pipeline consists of two steps: 1) Generate an initial pool of $N$ bottom-up proposals. Our method is agnostic to the precise bottom-up proposal method. The main point of this step is to prune out the obviously unlikely windows so that DeepBox can focus on the hard negatives. 2) Rerank the proposals using scores obtained by the DeepBox network. We rerank each proposal by cropping out the proposal box and feeding it into a CNN, as described by Girshick et al.~\cite{GirshickCVPR2014}. Because highly overlapping proposals are handled independently, this strategy is computationally wasteful and thus slow. A more sophisticated and much faster approach, using ideas from~\cite{HeECCV2014,FastRCNN} is described in Section~\ref{subsec:fast}.

%\subsection{Proposal Pool Preparation}
\label{subsec:eb}

For datasets without many small objects (e.g. PASCAL), we often only need to 
re-rank the top $~2000$ proposals to obtain good enough recall. As shown by ~\cite{ZitnickECCV2014}, increasing the number of Edge box proposals beyond $2000$ leads to only marginal increase in recall. For more challenging datasets with small objects (e.g. COCO~\cite{mscoco}), reranking more proposals continues to provide gains in recall beyond 2000 proposals.
\subsection{Network Architecture}\label{sec:net}

When it comes to the network architecture, we would expect that predicting the precise category of an object is harder than predicting objectness, and so we would want a simpler network for objectness. This also makes sense from a computational standpoint, since we do not want the object proposal scoring stage to be as expensive as the detector itself.

We organized our search for a suitable network architecture by starting from the architecture of~\cite{KrizhevskyNIPS2012} and gradually ablating it while trying to preserve performance. The ablation here can be performed by reducing the number of channels in the different layers (thus reducing the number of parameters in the layer),  by removing some layers, or by decreasing the input resolution so that the features computed become coarser.

The original architecture gave an AUC on PASCAL VOC of $0.76 (0.62)$ for IoU=0.5 (0.7). First, we changed the number of outputs of $fc6$ to $1024$ with other things fixed and found that performance remained unchanged. Then we adjusted the input image crop from $227\times227$ of the network to $120 \times 120$ and observed that the AUC dropped $1.5$ points for IoU=0.5 and $2.9$ points for IoU=0.7 on PASCAL. With this input size, we tried removing $fc6$ (drop: 2.3 points), $conv5$ (drop: 2.9 points), $conv5+conv4$ (drop: 10.6 points) and $conv5+conv4+conv3$ layers (drop: 6.7 points). This last experiment meant that  dropping all of $conv5, conv4$ and $conv3$ was \emph{better} than just dropping $conv5$ and $conv4$. This might be because $conv3, conv4$ and $conv5$ while adding to the capacity of the network are also likely overfit to the task of image classification (as described below, the convolutional layers are initialized from a model trained on ImageNet). We stuck to this architecture (i.e.,without $conv5$, $conv4$ and $conv3$) and explored different input sizes for the net. For an input size of $140\times140$, we obtained a competitive AUC of $0.74$ (for IoU=0.5) and $0.60$ (for IoU=0.7) on PASCAL, or equivalently a $4.4$ point drop against the baseline.

Our final architecture can be written down  as follows.
Denote by $conv(k,c,s)$ a convolutional layer with kernel size $k$, stride $s$ and number of output channels $c$. Similarly, $pool(k,s)$ denotes a pooling layer with kernel size $k$ and stride $s$, and $fc(c)$ a fully connected layer with $c$ outputs. Then, our network architecture is: \\ $conv(11,96,4)-pool(3,2)-conv(5,256,1)-fc(1024)-fc(2)$.\\ Each layer except the last is followed by a ReLU non-linearity. Our problem is a binary classification problem (object or not), so we only have two outputs which are passed through a softmax. Our input size is $140 \times 140$. 

While we finalized this architecture on PASCAL for Edge boxes, we show in Section~\ref{sec:experiments} that the same architecture works just as well for other datasets such as COCO~\cite{mscoco} and for other proposal methods such as Selective Search~\cite{UijlingsIJCV2013} or MCG~\cite{ArbelaezCVPR2014}.

\subsection{Sharing computation for faster reranking}
\label{subsec:fast}
Running the CNN separately on highly overlapping boxes wastes a lot of computation. He et al.~\cite{HeECCV2014} pointed out that the convolutional part of the network could be shared among all the proposals. Concretely, instead of cropping out individual boxes, we pass in the entire image into the network (at a high resolution). After passing through all the convolutional and pooling layers, the result is a feature map which is some fraction of the image size.  Given this feature map and a set of bounding boxes, we want to compute a fixed length vector for each box that we can then feed into the fully connected layers. To do this, note that each bounding box $B$ in the image space corresponds to a box $b$ in the feature space of the final convolutional layer, with the scale and aspect ratio of $b$ being dependent on $B$ and thus different for each box. He et al. propose to use a \emph{fixed} spatial pyramid grid to max-pool features for each box. While the size of the grid cell varies with the box, the number of bins don't, thus resulting in a fixed feature vector for each box. From here on, each box is handled separately. However, the shared convolutional feature maps means that we have saved a lot of computation.

One issue with this approach is that all the convolutional feature maps are computed at just one image scale, which may not be appropriate for all objects. He et al.~\cite{HeECCV2014} suggest a multiscale version where the feature maps are computed at a few fixed scales, and for each box we pick the best scale, which they define as the one where the area of the scaled box is closest to a predefined value. We experiment with both the single scale version and a multiscale version using three scales.

We use the implementation proposed by Girshick in Fast R-CNN~\cite{FastRCNN}. Fast R-CNN implements this pooling as a layer in the CNN (the RoI Pooling layer) allowing us to train the network end-to-end. 

To differentiate this version of DeepBox from the slower alternative based on cropping and warping, we call this version Fast DeepBox in the rest of the paper.

%We re-implement Fast DeepBox based on spatial pyramid pooling ~\cite{HeECCV2014} and Fast-RCNN ~\cite{FastRCNN} framework. Like Fast-RCNN, our pyramid has only one level. The same small network is used, with an ROI pooling layer inserted between conv layers and fc layers. 

%Here we describe the functions and significance of ROI pooling layer. E Let $r$ be of size $h \times w$, and located at $(x,y)$. Then, the ROI pooling layer takes $h \times w$ input block at $(x,y)$ and outputs $H \times W$ block by maxpooling over $H \times W$ output bins. The bin size ($\approx h/H \times w/W$) is adaptively chosen following ~\cite{HeECCV2014}. For different ROIs, this $H \times W$ is fixed because the fc layers on top only accept inputs with fixed dimension. ROI pooling offers a great advantage that, instead of warping every ROI and passing it into the net, we can pass in the whole image through once and sample ROIs from the top of conv layers. This shares significant amount of computation among ROIs that occur in conv layers, which is the computation bottleneck of deep nets.

%The net was trained as a two-class objectness detector following the same scheme as in ~\ref{subsubsec:sliding} and ~\ref{subsubsec:retrain}. Since we have a shallow network, we downsized the proposals by $0.125$ when it goes from the image space to the ROI-pooling layer. 

\subsection{Training Procedure}

\subsubsection{Initialization}
The first two convolutional layers were initialized using the publicly available Imagenet model \cite{KrizhevskyNIPS2012}. This model was pretrained on $1000$ Imagenet categories for the classification task. The $fc$ layers are initialized randomly from Gaussian distribution with $\sigma=0.01$. Our DeepBox training procedure consists of two stages. Similar to the classical notion of bootstrapping in object detection, we first train an initial model to distinguish between object boxes and randomly sampled sliding windows from the background. This teaches it the difference between objects and background. To enable it to do better at correcting the errors made by bottom-up proposal methods, we run a second training round where we train the model on bottom-up proposals from a method such as Edge boxes.
% \textcolor{red}{We found that this two stage training was important when working with Slow DeepBox but was not needed for Fast DeepBox. This discrepancy may be due to artifacts introduced by cropping and warping}.
%~\textcolor{red}{However, if training time is a concern, we found that training just on the bottom-up proposals will only slightly compromise the performance. Thus, the sliding window traing could be considered as an optional performance boost.}
 %In objectness generalization study, we separated the training (which includes Imagenet) and testing categories to show that true objectness can be learnt by a network. In object detection study, we trained on all categories and evaluated on all categories. 

\subsubsection{Training on Sliding Windows}
\label{subsubsec:sliding}
First we generate negative windows by simple raster scanning. The sliding window step size is selected based on the box-searching strategy of Edge boxes\cite{ZitnickECCV2014}. Following Zitnick et al~\cite{ZitnickECCV2014}, we use $\alpha$ to denote the IoU threshold for neighboring sliding windows, and set it to 0.65. We generated windows in $5$ aspect ratios: $(w:h)=(1:1)$, $(2:3)$, $(1:3)$, $(3:2)$, and $(3:1)$.  Negative windows which overlap with a ground truth object by more than $\beta_{-} = 0.5$ are discarded.

To obtain positives, we randomly perturb the corners of ground
truth bounding boxes. Suppose a ground truth bounding box has coordinates $(x_{min}, y_{min}, x_{max}, y_{max})$, with the width denoted by $w$ and the height denoted by $h$.
Then the perturbed coordinates are distributed as: 
\begin{eqnarray}
x'_{min}&\sim &\text{unif}(x_{min}-\gamma w,x_{min}+\gamma w)\\
y'_{min}&\sim &\text{unif}(y_{min}-\gamma h,y_{min}+\gamma h)\\
x'_{max}&\sim&\text{unif}(x_{max}-\gamma w,x_{max}+\gamma w)\\
y'_{max}&\sim&\text{unif}(y_{max}-\gamma h,y_{max}+\gamma h)
\end{eqnarray}
 where $\gamma=0.2$ defines the level of noise. Larger
$\gamma$ introduces more robustness into positive training samples, but might hurt localization. In practice we found that $\gamma=0.2$ works well. In case some perturbed points go out of the image, we set them to stay on the image border. Perturbed windows that overlap with ground truth boxes by less than $\beta_{+}=0.5$ are discarded.

%~\textcolor{red}{These two paragraphs about training details are removed, since sliding windows are not as important as before.}
%Finally, we balanced the ratio of positive and negative windows at $1:3$ at training time. Given any image, we first generate $N$ negative windows. If there are $g$ objects in the image, we generate $\lfloor \frac{N}{3g} \rfloor$ positive windows for each object.

%Following \cite{KrizhevskyNIPS2012}, momentum is set to 0.9 and weight decay to 0.0005. The initial learning rate is 0.001, which we decrease by 0.1 every 20,000 iterations. We train the network for 60,000 iterations with a mini-batch size of 128. 

\subsubsection{Training on Hard-negatives}
\label{subsubsec:retrain}
Next, we trained the net on bottom-up proposals from a method such as Edge boxes~\cite{ZitnickECCV2014}. Compared to sliding windows, these
proposals contain more edges, texture and parts of objects, and thus form hard negatives. While sliding windows trained the net to distinguish primarily between background and objects, training on these proposals allows the net to learn the notion of complete objects and better adapt itself to the errors made by the bottom-up proposer. 
 This is critical: without this stage, performance drops by $18\%$ on PASCAL from $0.60$ AUC at IoU=0.7 to $0.48$. 
%We believe objectness is a high-level concepts. Thus, to train the net on object-rich proposals is essential for the net to grasp the concept of complete
%objects.

The window preparation and training procedure is as described in Section~\ref{subsubsec:sliding}, except that sliding windows are replaced with Edge boxes proposals. We set the overlap threshold $\beta_{+}=0.7$ slightly higher,
so that the net can learn to distinguish between tight and loose bounding
boxes. We set $\beta_{-}=0.3$ below which the windows are labeled negative. Windows with IoU $\in[0.3,0.7]$ are discarded lest they confuse the net. 

We balanced the ratio of positive and negative windows at $1:3$ at training time for both stages. Momentum is set to 0.9 and weight decay to 0.0005. The initial learning rate is 0.001, which we decrease by 0.1 every 20,000 iterations. We train DeepBox for 60,000 iterations with a mini-batch size of 128. 

The Fast DeepBox network was trained for $120k$ iterations in both sliding window and hard negative training. Three scales $[400,600,900]$ were used for both training and testing.

%\subsection{Test time procedure}

%Given an image in test time, we first prepare the proposal pool following
%the approach described in Section~\ref{subsec:eb}. Since each proposal has a different aspect ratio, we warped every proposal to some fixed size before feeding
%it into the net. The net then assigned each box an objectness score, by which we re-ranked the proposals. Both training and testing was performed on a K40 GPU.

\section{Experiments}
\label{sec:experiments}
All experiments except those in Section~\ref{exp:fast} and \ref{exp:det} were done using DeepBox. We evaluate Fast DeepBox in Section~\ref{exp:fast}, and as a final experiment plug it into an object detection system in Section~\ref{exp:det}.
\subsection{Learning objectness on PASCAL and COCO}
\label{sec:pascal_coco_exp}
In our first set of experiments, we evaluated our approach on PASCAL VOC 2007~\cite{pascal-voc-2007} and on the newly released COCO~\cite{mscoco} dataset. For these experiments, we used Edge boxes~\cite{ZitnickECCV2014} for the initial pool of proposals. On PASCAL we reranked the top 2048 Edge box proposals, whereas on COCO we re-ranked all. We used the network architecture described in Section~\ref{sec:net}. For results on PASCAL VOC 2007, we trained our network on the trainval set and tested on the test set. For results on COCO, we trained our network on the train set and evaluated on the val set.

\subsection{Comparison to Edge boxes} 
\begin{figure}
\begin{center}
\includegraphics[width=0.9\columnwidth]{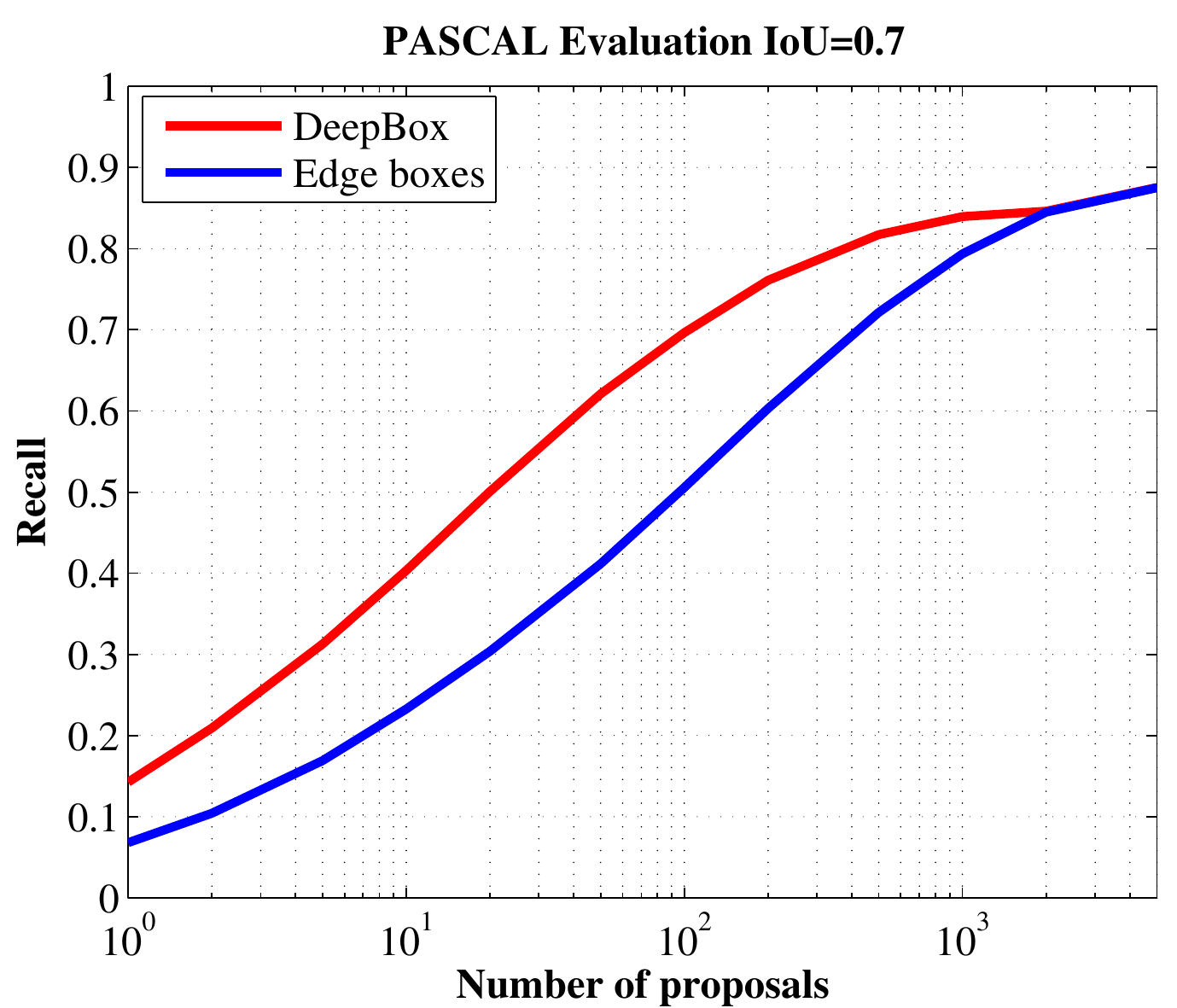}
\end{center}
\caption{PASCAL Evaluation IoU=0.7. DeepBox starts off much higher than Edge boxes. The wide margin continues all the way until $500$ proposals and gradually decays. The two curves join at $2048$ proposals because we chose to re-rank this number of proposals. }
\label{fig:voc07}
\end{figure}

\begin{figure}
\begin{center}
\includegraphics[width=0.9\columnwidth]{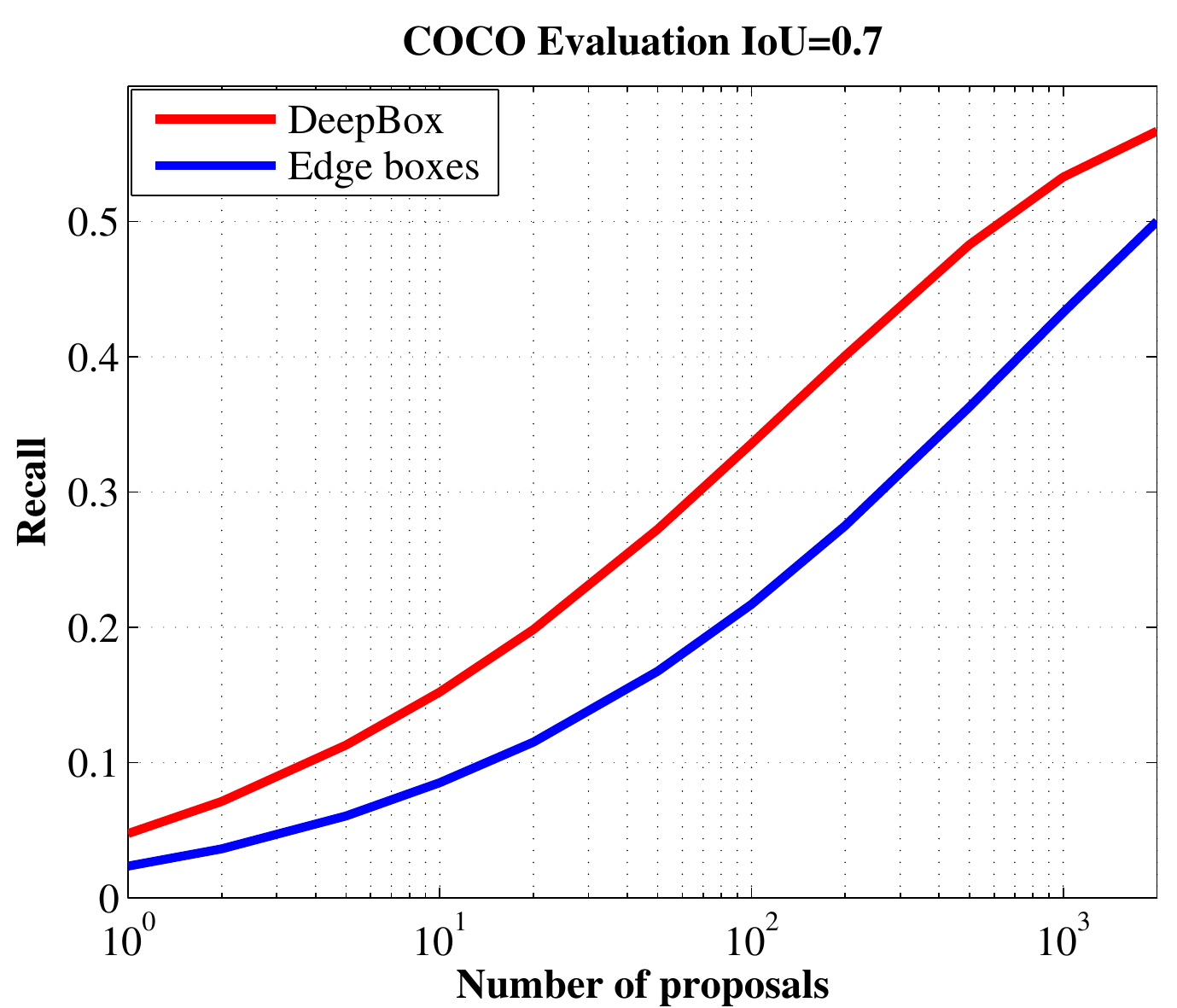}
\end{center}
\protect\caption{MS COCO Evaluation IoU=0.7. The strong gain demonstrated by DeepBox on COCO suggests that our learnt objectness is particularly helpful in a complicated dataset.}
\label{fig:coco07}
\end{figure}

We first compare our ranking to the ranking output by Edge boxes. Figure~\ref{fig:voc07} plots Recall vs Number of Proposals in PASCAL VOC 2007 for IoU=0.7\footnote{We computed recall using the code provided by~\cite{ZitnickECCV2014}}. We observe that DeepBox outperforms Edge boxes in all regimes, especially with a low number of proposals. The same is true for IoU=0.5 (not shown). The AUCs (Areas Under Curve) for DeepBox are $0.74(0.60)$ vs Edge box's $0.60(0.47)$ for IoU=0.5(0.7), suggesting that DeepBox proposals are $24\%$ better at IoU=0.5 and $26\%$ better at IoU=0.7 compared to Edge boxes.  Figure~\ref{fig:coco07} plots the same in COCO. The AUCs (Areas Under Curve) for DeepBox are $0.40(0.28)$ vs Edge boxes's $0.28(0.20)$ for IoU=0.5(0.7), suggesting DeepBox proposals are $40\% (43\%)$ better than Edge boxes. If we re-ranked top 2048 proposals instead, the AUCs are $0.38(0.27)$.

\begin{figure}
\begin{center}
\includegraphics[width=0.9\columnwidth]{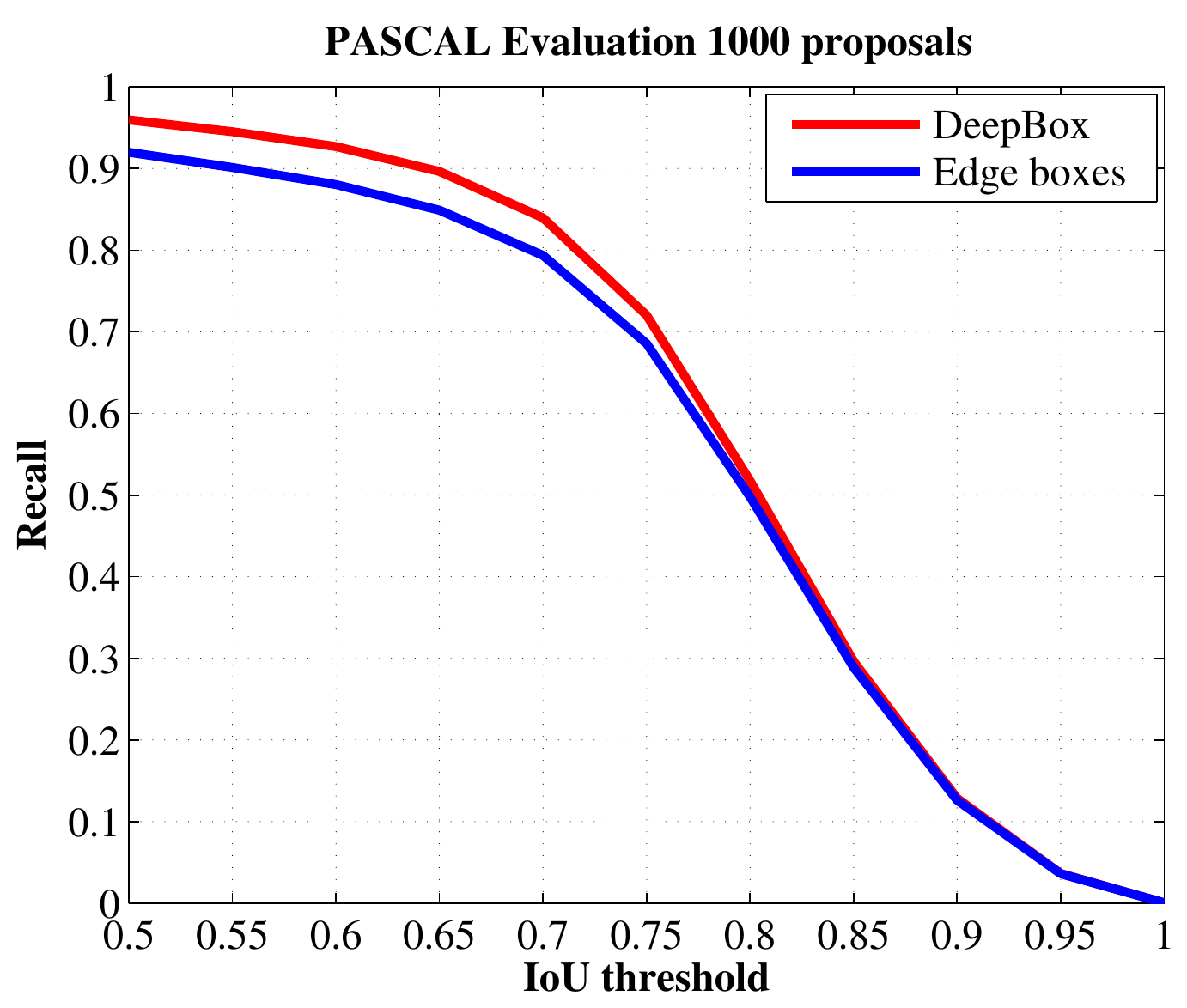}
\end{center}
\protect\protect\caption{Variation of recall with IoU threshold at $1000$ proposals. DeepBox (average recall: 57.0\%) is better than Edge boxes (average recall: 54.4\%) in all regimes. Comparisons to other proposal methods is shown in the supplementary. }
\label{fig:recvsiu}
\end{figure}

On PASCAL, we also plot the Recall vs IoU threshold curves at 1000 proposals in Figure~\ref{fig:recvsiu}. At 1000 proposals,
the gain of DeepBox is not as big as at 100-500 proposals, but we
still see that it is superior in all regimes of IoU. 

Part of this performance gain is due to small objects, defined as objects with area less than $2000$. On COCO the AUCs for DeepBox (trained on all categories) are $(0.161,0.080)$, while the numbers for Edge boxes are $(0.061,0.030)$ for IoU($0.5$,$0.7$). DeepBox outperforms Edge boxes by more than $160\%$ on small objects.

\paragraph{Comparison to other proposal methods} We can also compare our ranked set of proposals to all the other proposals in the literature. We show this comparison for IoU=0.7 in Table~\ref{tab:results}.
The numbers are obtained from \cite{ZitnickECCV2014} except for MCG and DeepBox which we computed ourselves. In Table~\ref{tab:results}, the metrics are the number of proposals needed to achieve 25\%, 50\% and 75\% recall and the maximum recall using 5000 boxes. 

\begin{table}
\begin{center}
{\small
\begin{tabular}{lcccccc}
\hline 
 &  AUC  &  25\%  &  50\%  &  75\%  &  Recall  &  Time\\
 & & & & & (\%) & (s) \\
 \hline
 BING\cite{MingCVPR2014}  & 0.20  &  292  &  -  &  -  &  29  &  \textbf{0.2}\\
 Ranta\cite{RantaCVPR2014}  &  0.23  &  184  &  584  &  -  &  68  &  10\\
 Objectness\cite{AlexeTPAMI2012}  &  0.27  &  27  &  -  &  -  &  39  &  3\\
 Rand. P.\cite{ManenICCV2013} &  0.35  &  42  &  349  &  3023  &  80  &  1\\
 Rahtu \cite{RahtuICCV2011} &  0.37  &  29  &  307  &  -  &  70  &  3\\
 SelSearch \cite{UijlingsIJCV2013} &  0.40  &  28  &  199  &  1434  &  \textbf{87}  &  10\\
 CPMC \cite{CarreiraECCV2012} &  0.41  &  15  &  111  &  -  &  65  &  250\\
 MCG \cite{ArbelaezCVPR2014} &  0.48  &  10  &  81  &  871 &  83  &  34\\
 E.B \cite{ZitnickECCV2014} &  0.47  &  12  &  108  &  800  &  \textbf{87}  &  0.25\\
 DeepBox  & \textbf{ 0.60} & \textbf{ 3} & \textbf{ 20} & \textbf{ 183} & \textbf{ 87}  &  2.5\\
\hline 
\end{tabular}
}
 \end{center}
\protect\protect\caption{Comparison of DeepBox with existing object proposal techniques. All numbers are for an IoU threshold of 0.7. The metrics
are the number of proposals needed to achieve 25\%, 50\% and 75\%
recall and the maximum recall using 5000 boxes. DeepBox
outperforms in all metrics. Note that the timing numbers for DeepBox include computation on the GPU.}

\label{tab:results} 
\end{table}

\subsection{Visualization of DeepBox proposals}
\label{sec:vis}
\begin{figure*}
\begin{center}
\includegraphics[width=0.84\textwidth]{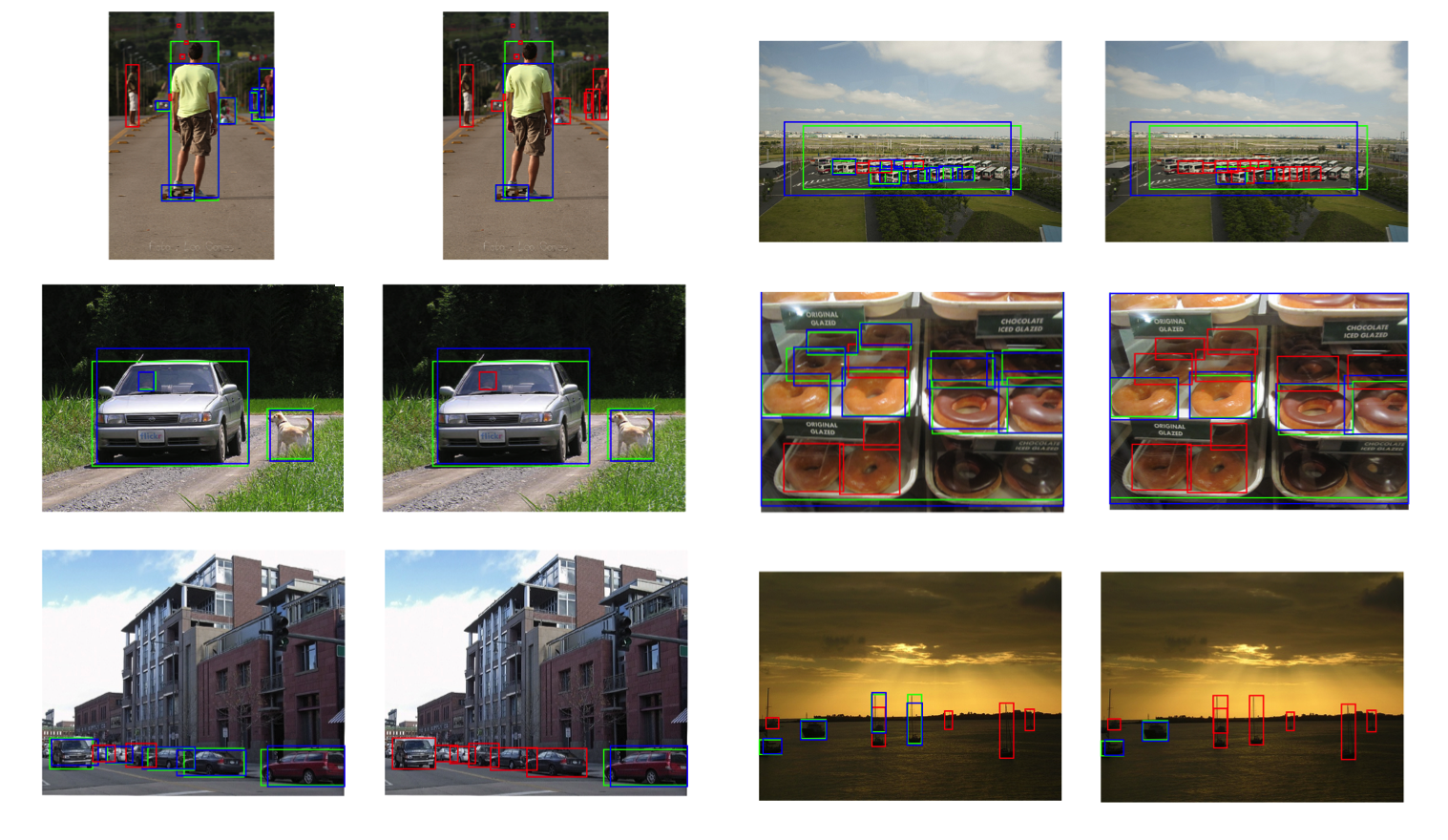}
\end{center}
\caption{Visualization of hits and misses. In each image, the green boxes are ground truth boxes for which a highly overlapping proposal exists (with the corresponding proposals shown as blue boxes) and red boxes are ground truth boxes that are missed. The IoU threshold is $0.7$. We evaluated 1000 proposals per image for COCO and 500 proposals per image for PASCAL. The left image in every pair shows the result of  DeepBox ranking, and the right image shows the ranking from Edge boxes. In cluttered scenes, DeepBox has a higher recall. See Section~\ref{sec:vis} for a detailed discussion.}
\label{fig:vis1}
\end{figure*}

\begin{figure*}
\begin{center}
\includegraphics[width=1\textwidth]{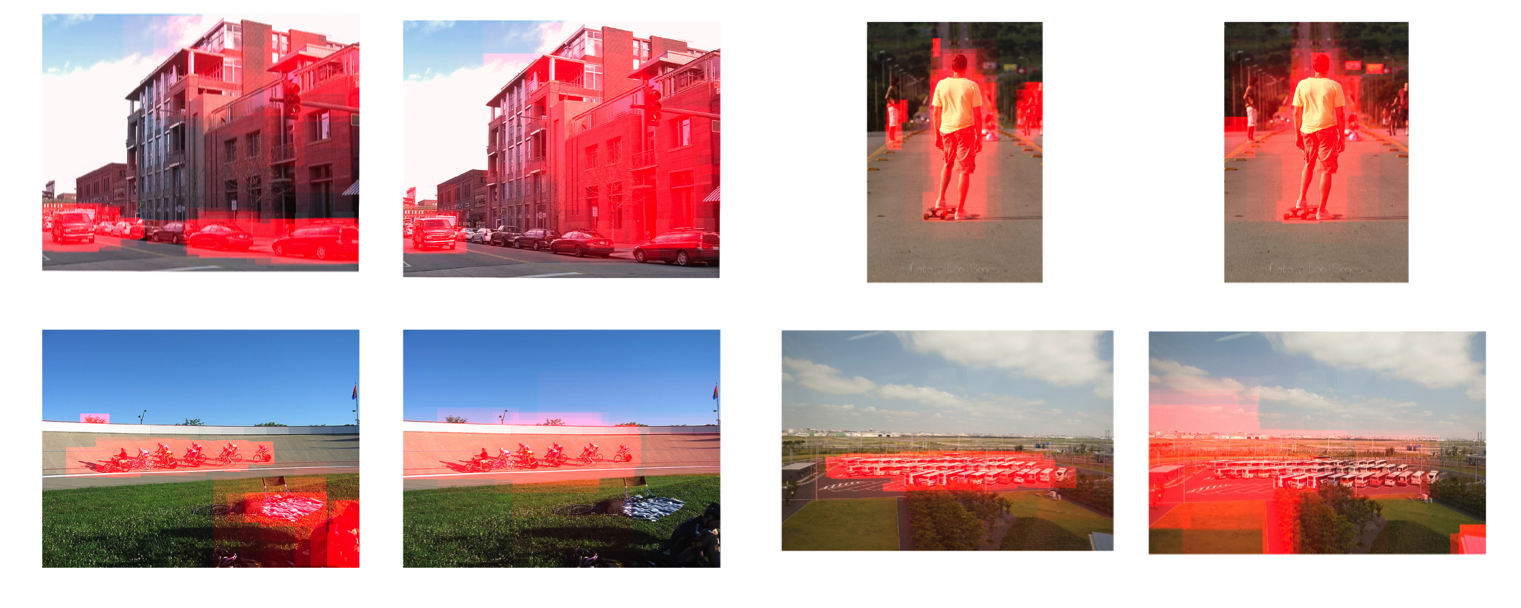}
\end{center}
\protect\caption{In each image we show the distribution of the proposals produced by pasting a red box for each proposal. Only the top 100 proposals are shown. For each pair of images, DeepBox is on the left and Edge boxes is the right. DeepBox proposals are more tightly concentrated on the objects. See Section~\ref{sec:vis} for a detailed discussion. }
\label{fig:vis2}
\end{figure*}

Figures~\ref{fig:vis1} and ~\ref{fig:vis2} visualizes DeepBox and Edge box performance on PASCAL and COCO images. Figure~\ref{fig:vis1} shows the ground truth boxes that are detected (``hits", shown in green, with the corresponding best overlapping proposal shown in blue) and those that are missed (shown in red) for both proposal rankings. We observe that in complicated scenes with multiple small objects or cluttered background, DeepBox significantly outperforms Edge box. The tiny boats, the cars parked by the road, the donuts and people in the shade are all correctly captured. 

This is also validated by looking at the distribution of
top 100 proposals (Figure~\ref{fig:vis2}), which is shown in red. In general, DeepBox's bounding boxes are very densely focused on the objects of interest while Edge boxes primarily recognize contours and often spread evenly across the image in a complicated scene.

\subsection{Generalization to unseen categories}

The high recall our method achieves on the PASCAL 2007 test set does not guarantee that our net
is truly learning general objectness. It is possible that the net
is learning just the union of 20 object categories and using that knowledge
to rank proposals. To evaluate whether the net is indeed learning a more general notion of objectness that extends beyond the categories it has seen during training, we did the following experiment:
\begin{itemize}
\item We identified the 36 overlapping categories between Imagenet and COCO.
\item We trained the net just on these overlapping categories on COCO with initialization from Imagenet. This means that during training, only boxes that overlapped highly with ground truth from the 36 overlapping categories were labeled positives, and others were labeled negatives. Also, when sampling positives, only the ground truth boxes corresponding to these overlapping categories were used to produce perturbed positives. This is equivalent to training on a dataset where all the other categories have not been labeled at all.
\item We then evaluated our performance on the rest of the categories in COCO (44 in number). This means that at test time, only proposed boxes that overlapped with ground truth from the other 44 categories were considered true positives. Again, this corresponds to evaluating on a dataset where the 36 training categories have not been labeled at all.
\end{itemize}

All other experimental settings remain the same. As before, we use Edge boxes for the initial pool of proposals which we rerank, and compare the ranking we obtain to the ranking output by edge boxes.

\begin{figure}
\begin{center}
\includegraphics[width=0.9\columnwidth]{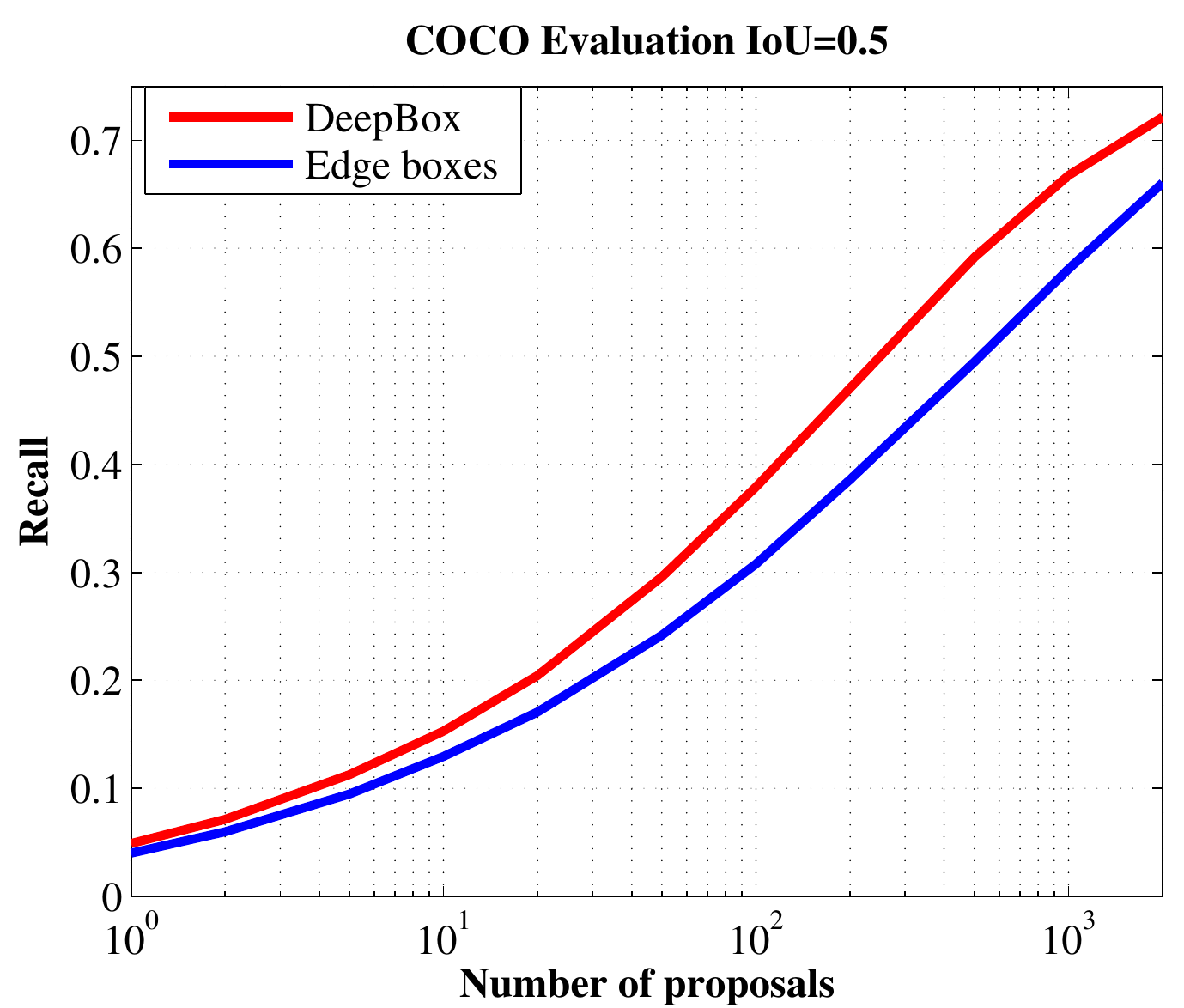}
\end{center}
\protect\caption{Evaluation on unseen categories, when ranking all proposals, at IoU=0.5.}
\label{fig:unseen_coco_05}
\end{figure}

When reranking the top 2048 proposals, DeepBox achieved $15.7\% (15.3\%)$ AUC improvement over Edge boxes for IoU=0.5 (0.7). When reranking \emph{all} proposals output by Edge boxes, DeepBox achieved $18.5\% (16.2\%)$ improvement over Edge boxes for IoU=0.5 (0.7) (Figure~\ref{fig:unseen_coco_05}). In this setting, DeepBox outperforms Edge box in all regimes on unseen categories for both IoUs. 

\begin{figure*}
\begin{center}
\includegraphics[width=.85\textwidth]{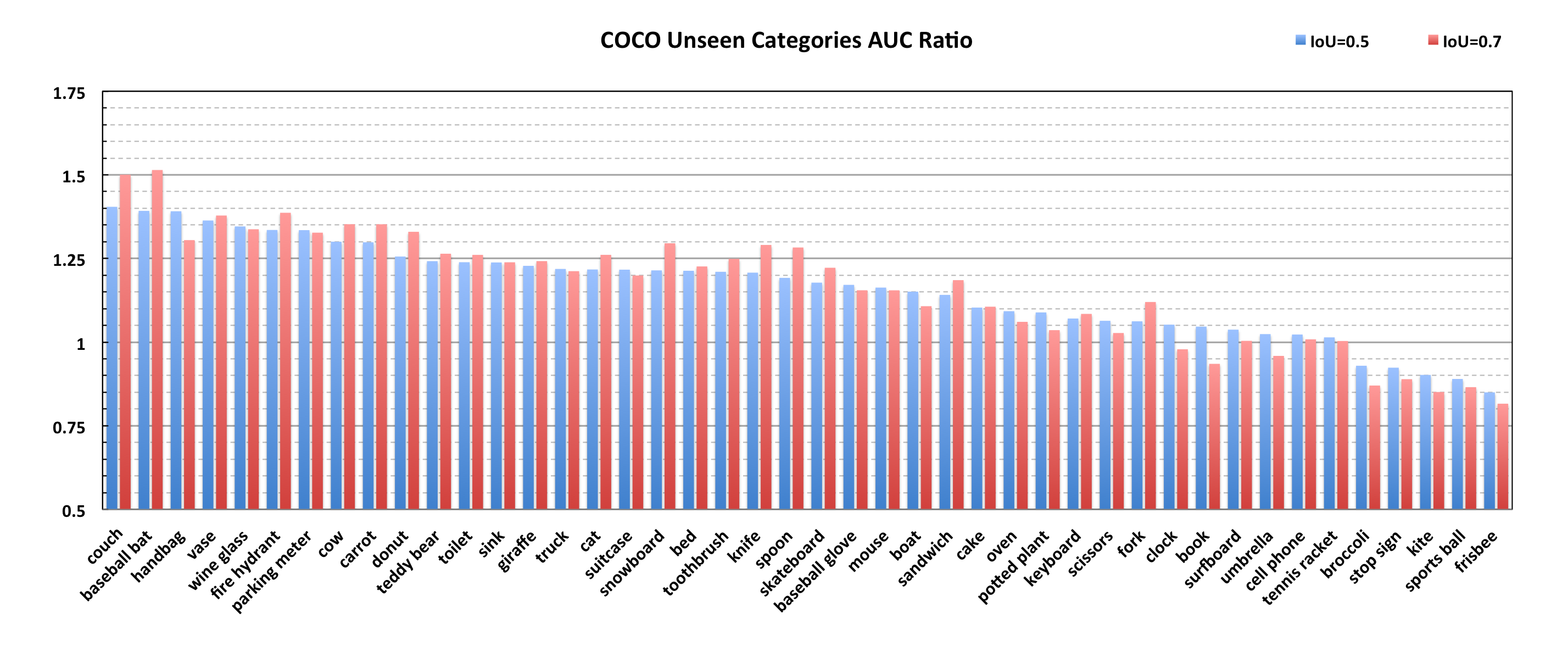}
\end{center}
\caption{Evaluation on unseen categories: category-wise breakdown. This demonstrates that DeepBox network has learnt
a notion of objectness that generalizes beyond training categories.}
\label{fig:unseen_ratio}
\end{figure*}

In Figure~\ref{fig:unseen_ratio}, we plot the ratio of the AUC obtained by  DeepBox to that obtained by Edge box for all the 44 testing categories. In more than half of the testing categories, we obtain an improvement greater than $20\%$, suggesting that the gains provided by DeepBox are spread over multiple categories. This suggests that DeepBox is actually learning a class-agnostic notion of objectness. Note that DeepBox performs especially well for the animal super category because all animal categories have similar geometric structure and training the net on some animals helps it recognize other animals. This validates our intuition that there are high-level semantic cues that can help with objectness. In the sports super category, DeepBox performs worse than Edge boxes, perhaps because most objects of this category have salient contours that favor the Edge boxes algorithm. 

These results on COCO demonstrate that our network has learnt
a notion of objectness that generalizes beyond training categories.
%This knowledge enables DeepBox to produce more accurate proposals
%than Edgebox and other proposers in Table 1.

\subsection{DeepBox using Large Networks}

We also experimented with larger networks such as ``Alexnet"~\cite{KrizhevskyNIPS2012} and ``VGG"~\cite{SimonyanZ14a}. Unsurprisingly, large networks capture objectness better than our small network. However, the difference is quite small:  With VGG, the AUCs on PASCAL are $0.78(0.65)$ for IoU=0.5(0.7). The numbers for Alexnet are $0.76(0.62)$. In comparison our network achieves $0.74(0.60)$. 

For evaluation on COCO, we randomly selected 5000 images and computed AUCs using the  VGG and Alexnet architecture. All networks re-rank just top 2048 Edge box proposals. At IoU=0.5, VGG gets $0.43$ and Alexnet gets $0.42$, compared to the $0.38$ obtained by our network.  At IoU=0.7, VGG gets $0.31$ and Alexnet gets $0.30$, compared to the $0.27$ obtained by our network. When re-ranking all proposals, our small net gets $0.40(0.28)$ for IoU=0.5(0.7).

These experiments suggest that our network architecture is sufficient to capture the semantics of objectness, while also being much more efficient to evaluate compared to the more massive VGG and Alexnet architectures.

\subsection{DeepBox with Other Proposers}

It is a natural question to ask whether DeepBox framework applies to other bottom-up proposers as well. We experimented with MCG and Selective Search by just re-ranking top 2048 proposals. (For computational reasons, we did this experiment on a smaller set of 5000 COCO images.) We experimented with two kinds of DeepBox models: a single model trained on Edge boxes, and separate models trained on each proposal method (Table ~\ref{tab:obj_props}). 

A model trained on Edge boxes does not provide much gains, and indeed sometimes hurts, when run on top of Selective Search or MCG proposals. However, if we retrain DeepBox separately on each proposal method, this effect goes away. In particular, we get large gains on Selective Search for both IoU thresholds (as with Edge boxes). On MCG, DeepBox does not hurt, but does not help much either. Nevertheless, the gains we get on Edge boxes and Selective Search suggest that our approach is general and can work with any proposal method, or even ensembles of proposal methods (a possibility we leave for future work).

\begin{table}
\begin{center}
{\small
\begin{tabular}{lcccccc}
\hline 
 &  Vanilla  &  DeepBox   &  DeepBox  &  Total\\
 & & (Trained on & (Finetuned) & Time \\
 & & ~\cite{ZitnickECCV2014}) &  & (s)\\
 \hline
 Sel. Se.  &  0.27/0.17  &  0.27/0.19  &  0.32/0.22  & 12.5 \\
 MCG  &  \textbf{0.38}/\textbf{0.25}  &  0.30/0.22  &  0.37/0.26  & 36.5 \\
 Edge Box &0.28/0.20 &  \textbf{0.38}/\textbf{0.27}  &  \textbf{0.38}/\textbf{0.27}  &  \textbf{2.5}\\
\hline 
\end{tabular}
}
\end{center}
\protect\protect\caption{DeepBox on top of other proposers. For each method we show the AUC at IoU=0.5/0.7 of (left to right) the original ranking, the reranking produced by DeepBox trained on Edge boxes, and that produced after finetuning on the corresponding proposals.}
%We show in order the method on its own, the method with a DeepBox ranker trained on Edge boxes,  the finetuned on each method separately. The two numbers are the AUC at IoU=0.5/0.7. Time taken includes the DeepBox ranker.}
\label{tab:obj_props} 
\end{table}

\subsection{Fast DeepBox}
\label{exp:fast}
%Fast-RCNN ~\cite{FastRCNN} code base was used for both training and testing. 
We experimented with Fast DeepBox on COCO.
The AUCs for multiscale Fast DeepBox are $0.40(0.29)$ vs Edge box's $0.28(0.20)$ for IoU=0.5(0.7), a gain of $41\% (44\%)$ over Edge boxes. When we re-ranked top 2048 proposals instead, the AUCs are $0.37(0.27)$. Compared with DeepBox's multi-thread runtime of $2.5s$, Fast DeepBox (multiscale) is an order of magnitude faster: it takes $0.26$s to re-rank all proposals or $0.12$s to re-rank the top-$2000$. This compares favorably to other bottom-up proposals. In terms of average recall with 1000 proposals, our performance (0.39) is better than GOP (0.36)~\cite{KrahenbuhlECCV2014}, Selective Search (0.36)~\cite{UijlingsIJCV2013} and Edge boxes (0.34)~\cite{ZitnickECCV2014}, and is about the same as MCG (0.40)~\cite{ArbelaezCVPR2014} while being almost 70 times faster. In contrast, Deepmask achieves 0.45 with a much deeper network at the expense of being 3 times slower (1.6 s) ~\cite{Deepmask15}.

With a small decrease in performance, Fast DeepBox can be made much faster. With a single scale, AUC drops by about $0.005$ when re-ranking top-$2000$ proposals and $0.01$ when re-ranking all proposals. However, it only takes $0.11$s to re-rank all proposals or $0.060$s for the top-$2000$. One can also make training faster by removing the scanning-window stage and using a single scale. This speedup comes with a drop in performance of $0.015$ when reranking all proposals compared to the multiscale two-stage version.

%On the other hand, single-stage single-scale training on Edge boxes proposals work reasonably well with AUCs $0.38(0.27)$ on COCO. For each image, single-scale Fast DeepBox takes $0.11$s to re-rank all proposals or $0.060$s to re-rank the top-$2000$ proposals. This might be the right choice if computational complexity is an issue.

\subsection{Impact on Object Detection}
\label{exp:det}
The final metric for any proposal method is its impact on object detection. Good proposals not only reduce the computational complexity but can also make object detection easier by reducing the number of candidates that the detector has to choose from~\cite{HosangArxiv2015,FastRCNN}.  We found that this is indeed true: when using 500 DeepBox proposals, Fast-RCNN (with the VGG-16 network) gives a mAP of \textbf{37.8}\% on COCO Test at IoU=0.5, compared to only \textbf{33.3}\% when using 500 Edge Box proposals. Even when using 2000 Edge Box proposals, the mAP is still lower (35.9\%). For comparison, Fast R-CNN using 2000 Selective Search proposals gets a mean AP of 35.8\%, indicating that with just 500 DeepBox proposals we get a 2 point jump in performance.
%We used Fast DeepBox proposals in a Fast R-CNN~\cite{FastRCNN} object detection pipeline, and compared it to the performance obtained by using the same number of Edge boxes. On COCO, with the VGG-16 network we achieved ~\textbf{37.8}\% mAP at IoU = 0.5 using top 500 fast DeepBox proposals, compared with Fast R-CNN's $35.9$ mAP in COCO leader board. %In addition, we augment the table given by DeepMask ~\cite{Deepmask15} to benchmark Fast-DeepBox with the state-of-the-art techniques ~\ref{tab:deepmask}. (~\textbf{TBD for all results}) 

%\begin{table}
%\begin{center}
%{\small
%\begin{tabular}{lcccccc}
%\hline 
% &  AR@ &  AR@  &  AR@  &  AUC  &  Time  & mAP\\
% & 10 & 100 & 1000 &  & (s) &  \\
% \hline 
% Sel. S. \cite{UijlingsIJCV2013} &  .05  &  .16  &  .36  &  .13 &  10  &  35.8\\
% MCG  \cite{ArbelaezCVPR2014} &  .10  &  .25  &  .40 & .18 &  34  &  NA\\
% E.B. \cite{ZitnickECCV2014} &  .07  &  .18  &  .34  &  .14  &  .25  &  NA\\
% GOP \cite{KrahenbuhlECCV2014} &  .04  &  .18  &  .36 &  .13  &  1.0  &  NA\\
% D. Box  & .12 & .25 & .39 & .19 & .50 & 37.8\\
% D. Mask \cite{Deepmask15}  & .15 & .31 & .45 & .23 & 1.6  & NA\\
%\hline 
%\end{tabular}
%}
% \end{center}
%\protect\protect\caption{Comparison of DeepBox with state-of-the-art object proposers on COCO.}
%
%\label{tab:deepmask} 
%\end{table}

\section{Discussion and Conclusion}
 We have presented an efficient CNN architecture that learns a semantic notion of objectness that generalizes to unseen categories. We conclude by discussing other applications of our objectness model.

First, as the number of object categories increases, the computational complexity of the detector increases and it becomes more and more useful to have a generic objectness system to reduce the number of locations the detector looks at. Objectness can also help take the burden of localization off the detector, which then has an easier task.

Second, AI agents navigating the world cannot expect to be trained on labeled data like COCO for every object category they see. For some categories the agent will have to collect data and build detectors on the fly. In this case, objectness allows the agent to pick a few candidate locations in a scene that look like objects and track them over time, thus collecting data for training a detector. Objectness can thus be useful for \emph{object discovery}~\cite{KangECCV2012}, especially when it captures semantic properties as in our approach.
%For such an application, a notion of objectness that is just based on bottom-up grouping cues might be insufficient. On the contrary, our model is more semantic, and as shown by our performance on unseen categories, is significantly better at localizing objects it has never seen before. 

%Another place where a trained objectness model can help is in the context of producing detection systems for a very large number of categories (say, a few million). Tricks such as SPP~\cite{HeECCV2014} can help make detection faster, but the final classification still scales linearly with the number of locations and the number of categories. Using our objectness model to drastically reduce the number of locations to be evaluated can help a lot. This can also take the burden of localization off the detector, which then has an easier task and can presumably be trained much more easily or with less training data. Again, these benefits will be apparent when we have a large number of classes.  

\section{Acknowledgement}
This work is supported by a Berkeley Graduate Fellowship and a Microsoft Research Fellowship. We thank NVIDIA for giving GPUs through their academic program too.

{\bibliographystyle{ieee}
\bibliography{egbib}
 } 
\end{document}